\documentclass[a4paper,twoside]{article}

\usepackage[pagebackref=false,breaklinks=true,colorlinks,bookmarks=false]{hyperref}

\usepackage{subcaption}
\usepackage{calc}
\usepackage{amssymb}
\usepackage{amstext}
\usepackage{amsmath}
\usepackage{amsthm}
\usepackage{multicol}
\usepackage{pslatex}
\usepackage{apalike}
\usepackage[bottom]{footmisc}

\usepackage{comment}

\usepackage{algorithm}
\usepackage[noEnd,indLines]{algpseudocodex}

\usepackage{bbm}

\usepackage{SCITEPRESS}     

\renewcommand{\R}{\mathbb{R}}

\newcommand{\e}{\boldsymbol{e}}
\newcommand{\et}{\textit{et al.}}
\newcommand{\tb}{\textbf}

\begin{document}

\title{Action tube generation by person query matching\\ for spatio-temporal action detection}

\author{\authorname{%
Kazuki Omi,
Jion Oshima, 
and
Toru Tamaki
\orcidAuthor{0000-0001-9712-7777}}
\affiliation{%
Nagoya Institute of Technology,
Japan}
\email{
\{k.omi.646,j.oshima.204\}@nitech.jp,
tamaki.toru@nitech.ac.jp}
}


\abstract{

This paper proposes a method for spatio-temporal action detection (STAD) that directly generates action tubes from the original video without relying on post-processing steps such as IoU-based linking and clip splitting. Our approach applies query-based detection (DETR) to each frame and matches DETR queries to link the same person across frames. We introduce the Query Matching Module (QMM), which uses metric learning to bring queries for the same person closer together across frames compared to queries for different people. Action classes are predicted using the sequence of queries obtained from QMM matching, allowing for variable-length inputs from videos longer than a single clip. Experimental results on JHMDB, UCF101-24, and AVA datasets demonstrate that our method performs well for large position changes of people while offering superior computational efficiency and lower resource requirements.

}

\onecolumn \maketitle \normalsize \setcounter{footnote}{0} \vfill

\section{Introduction}

In recent years, the importance of not only image recognition, but also video recognition, especially the recognition of human actions, has been increasing in various practical applications. Among the tasks that recognize human actions in videos, \emph{spatial-temporal action detection} (or \emph{STAD}) \cite{Chen_ICCV2023_EVAD,Sun_ECCV2018_ACRN,Wu_CVPR2019_LFB,Chen_ICCV2021_WOO,Yixuan_ECCV2020,Zhao_CVPR2022_TubeR,gritsenko_arXiv2023_STAR}, which detects the class, location, and interval of actions occurring in the video, is important in practical applications \cite{Ahmed_TITS2020}.

The goal of STAD is to create \emph{action tubes} (or simply \emph{tubes}) \cite{Gkioxari_CVPR2015_Tube}.
A tube is a sequence of bounding boxes of the same action class for the same person across the frame of the video.
Here, a \emph{bounding box} (or \emph{bbox}) indicates the rectangle enclosing the actor in the frame, and
a \emph{tubelet} links bounding boxes of the same action class for the same person within a video clip, a short segment of the original video consisting of several (8 or 16) frames, due to the manageability for video recognition models.
A \emph{tube}, on the other hand, links these boxes throughout the video. Thus, a tube provides spatio-temporal information for understanding the action's temporal progression over the frames and its spatial location within the frames.

\begin{figure}[t]
    \centering

    \begin{minipage}[b]{\linewidth}
        \centering
        \includegraphics[width=.8\linewidth]{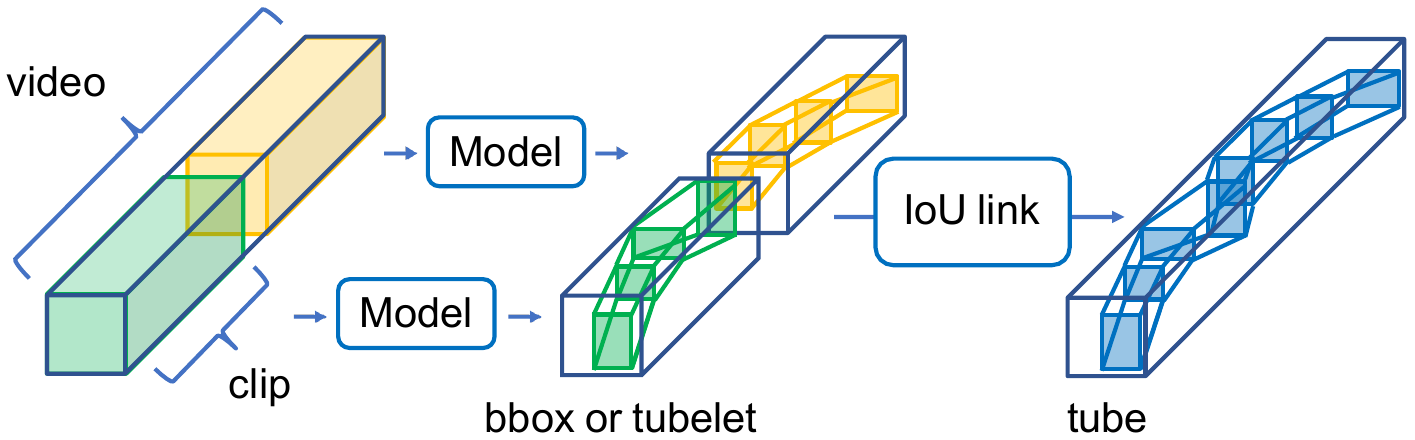}
    \subcaption{}
    \label{fig:method_conventional}
    \end{minipage}
    
    \medskip

    \begin{minipage}[b]{\linewidth}
        \centering
        \includegraphics[width=0.6\linewidth]{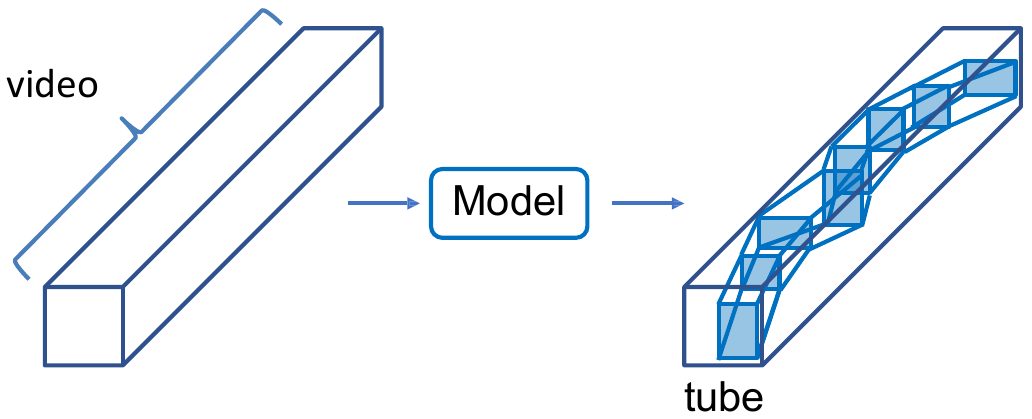}
    \subcaption{}
    \label{fig:method_proposed}
    \end{minipage}

    \caption{How to create action tubes.
    (a) A typical process of prior works. First, the original video is divided into multiple clips. Next, the clips are input into the model, which outputs \emph{bounding boxes} or \emph{tubelets}. Finally, these outputs are linked using IoU to create tubes.
    (b) The proposed method directly outputs \emph{tubes} of the original video.
    }
    
    \label{fig:methods}

\end{figure}

Although the length of the clips and the architecture of the models varies,
most prior works
\cite{köpüklü_arXiv2021_YOWO,Chen_ICCV2021_WOO,Kalogeiton_ICCV2017_ACT-Detector,Zhao_CVPR2022_TubeR,gritsenko_arXiv2023_STAR}
create tubes through the following post-processing (see Figure \ref{fig:method_conventional}). First, the original video is divided into multiple clips. Next, the model outputs a bounding box (bbox) or tubelet from the clip input. The model outputs are then linked using a linking algorithm \cite{Singh_ICCV2017_UCF101-24,Yixuan_ECCV2020} based on Intersection-over-Union (IoU) to create tubes.

However, there are two disadvantages to the post-processing of creating a tube by linking bbox or tubelets.
The first is IoU-based linking. It cannot handle large or fast motion of actors, and significant movements due to rapid camera motion or low fps \cite{Singh_WACV2023_TAAD}. For example, in actions with large fast movements such as ``diving'' and ``surfing,'' the IoU with adjacent frames may be small. In cases where the camera moves significantly, such as with in-car cameras rather than fixed cameras like surveillance cameras, even actions with small movements may have large displacements between frames, resulting in a small IoU. Thus, the types of actions and camera environments that can be linked by IoU are limited.

The second issue involves splitting a video into video clips.
Regardless of the video's context,
clips that are trimmed to a predetermined length may not be suitable for action recognition.
For example, a clip that only captures the first half of a ``jump'' action
might actually show the same movements as a ``crouch'' action,
but still needs to be identified as the ``jump'' action that will follow.
Recognizing actions from clips with incomplete action might lead to poor performance
and may also make training more difficult.

\begin{figure}[t]
    \centering

    \includegraphics[width=.7\linewidth]{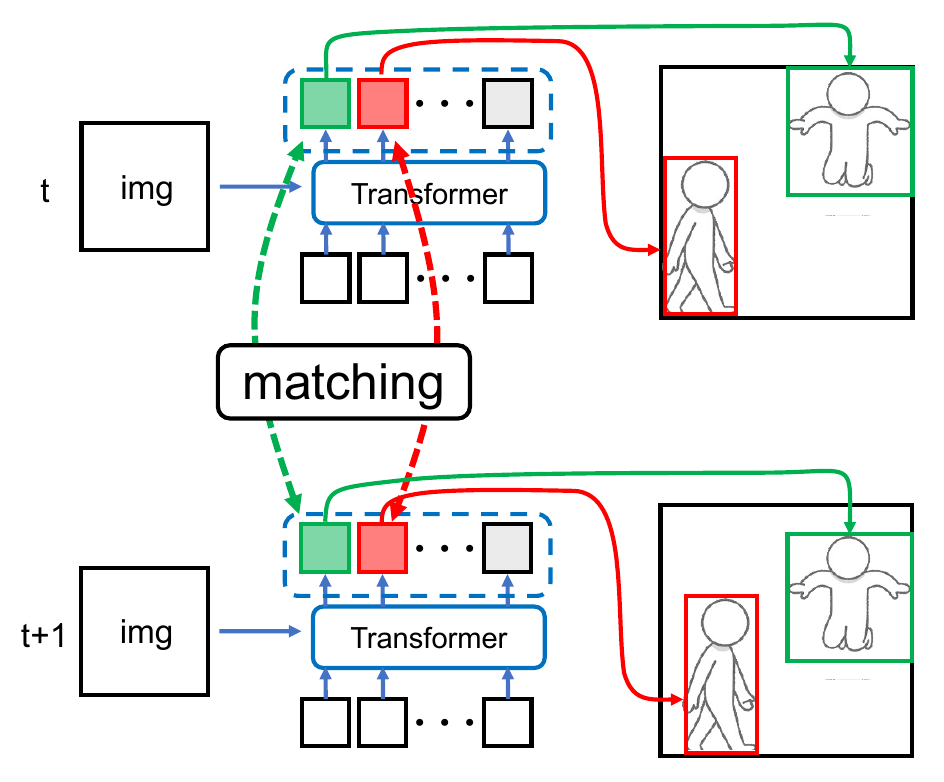}

    \caption{The proposed approach. Linking is performed by matching queries assigned to the same person, eliminating the need for the IoU-based linking.}

    \label{fig:approach}
\end{figure}

In this paper, we propose a method in which the model directly outputs tubes of the original video without using IoU-based linking and clip splitting. Although the goal of STAD is to create tubes, much attention has not been paid to models that directly output tubes. The proposed method eliminates post-processing because the model directly outputs tubes at the inference stage. Instead, the proposed method links the same person using query features. Specifically, inspired by the recent success of DETR \cite{Carion_ECCV2020_DETR}, we adopt an approach that applies query-based detection for each frame and matches queries to find the same person across frames. Our approach is illustrated in Figure \ref{fig:approach}. By linking with queries, we eliminate the need for IoU-based linking, allowing the detection of actions even with large displacements. Furthermore, by predicting action classes using a sequence of the queries obtained from matching, our method facilitates inputs of variable lengths for action recognition of the entire video.

\section{Related work}

\subsection{Object Detection}

The development of STAD aligns with advancements in object detection. In the past, there have been extensive studies on \emph{two-stage} methods that perform region detection and class identification in two stages, such as Fast-RCNN \cite{Girshick_ICCF2015_Fast_R-CNN}, and anchor-based \emph{one-stage} methods that perform them in one stage \cite{Redmon_CVPR2016_YOLO,Redmon_arXiv2017_YOLOv3,He_ICCV2017_Mask_R-CNN,Tan_CVPR2020_EfficientDet}. In recent years, with the introduction of DETR \cite{Carion_ECCV2020_DETR}, which formulates object detection as a bipartite matching problem between queries and ground truth bounding boxes, many \emph{query-based} detection methods \cite{zhu_arXiv2020_DeformableDETR,zhang_arXiv2022_DINO,Zong_ICCV2023_Co-DETR} have been proposed.

\subsection{Spatio-Temporal Action Detection}

Like object detection, STAD methods are generally categorized into \emph{two-stage} models \cite{Wu_CVPR2019_LFB,Feichtenhofer_2019ICCV_SlowFast,Pan_CVPR2021_ACAR,Fan_ICCV2021_MViT,Wu_CVPR2022_MeMViT}, which perform person detection and classification of actions in two steps, and \emph{one-stage} models \cite{Sun_ECCV2018_ACRN,Chen_ICCV2021_WOO,köpüklü_arXiv2021_YOWO,Zhao_CVPR2022_TubeR,gritsenko_arXiv2023_STAR}, which perform these tasks simultaneously.

Although most discussions in the literature emphasize the number of stages, this paper focuses on the type of model output. Specifically, we examine three types of model output: \emph{bounding box} (bbox), \emph{tubelet}, and \emph{tube}. This emphasis is important because, despite the objective of STAD being the prediction of action tubes, many prior works do not generate tube output.

\subsubsection{Bounding box}

Two types of models that output bounding box regions and corresponding action classes have been proposed;
one is to process frame-by-frame \cite{Weinzaepfel_ICCV2015,saha_arXiv2016_OJLA,Singh_ICCV2017_UCF101-24}, and the other is to take a clip as input for video feature extraction to output bounding box only for keyframes (such as the first or middle frame of the clip) \cite{köpüklü_arXiv2021_YOWO,Chen_ICCV2021_WOO,Feichtenhofer_2019ICCV_SlowFast,Feichtenhofer_2020CVPR_X3D,Sun_ECCV2018_ACRN}.

These methods temporally stride the input frame or input clip by one frame, obtaining bounding boxes for all frames, and then link the bounding boxes using a linking algorithm based on IoU \cite{Singh_ICCV2017_UCF101-24,Yixuan_ECCV2020} to create tubes.
Therefore, there are disadvantages due to IoU-based linking and the use of video clips.

\subsubsection{Tubelet}

Recently, models that output tubelets for a given clip end-to-end have also been proposed \cite{Kalogeiton_ICCV2017_ACT-Detector,Zhao_CVPR2022_TubeR,gritsenko_arXiv2023_STAR}. 3D cuboid-based approach \cite{Kalogeiton_ICCV2017_ACT-Detector,Yang_CVPR2019_STEP} is a representative approach that extends the 2D spatial anchors in the temporal direction. Cuboid-based methods assume that the temporal displacement of the actor is small and therefore are difficult to represent the dynamics of the tubelets \cite{Yixuan_ECCV2020}. TubeR \cite{Zhao_CVPR2022_TubeR} uses tubelet queries, which extend the object queries used in DETR, to express the tubelet dynamics. This is a query-based method, where one tubelet query is responsible for one tubelet within a clip, and it applies matching loss to learn to output tubelets that are robust to temporal displacement. Furthermore, query-based STAR \cite{gritsenko_arXiv2023_STAR} eliminates the need for an offline computed external memory bank \cite{Wu_CVPR2019_LFB}.

However, these methods still require the linking of the model output tubelets into tubes using IoU, thus retaining the drawbacks associated with IoU-based linking and the usage of clips.

\subsubsection{Tube}

There are few methods in which the model directly outputs a tube. In T-CNN \cite{Hou_ICCV2017_T-CNN}, tubelet proposals are generated for each clip, and tubes are created by linking these proposed tubelets using their action scores and overlaps. The created tube is then classified to obtain the final action score. Therefore, it has disadvantages of linking using IoU and using clips. On the other hand, TAAD \cite{Singh_WACV2023_TAAD} uses a tracking model with person features to link the same person, and the model directly outputs the tube. Hence, it does not have the disadvantages of IoU linking or clips. However, this method is not query-based and only uses the features of the regions obtained by tracking, resulting in the loss of information about the appearance of objects around the actor \cite{Sun_CVPR2021_Sparse_R-CNN}. Also, it does not consider cases where the detected person is not performing an action or performing actions sequentially.

Meanwhile, the proposed method is query-based and does not require linking using IoU. Therefore, it can detect actions with large displacements and classifies actions using the appearance information of surrounding objects and the global scene context. Additionally, the proposed method considers cases where the detected person is not performing an action.

\subsection{Object Tracking}

The proposed method uses a tracking approach inspired by multiple object tracking (MOT) tasks to link the same person using visual features. Tracking approaches are categorized by the type of cues used to link the same person. The first approach is based on IoU. SORT \cite{bewley_ICIP2016_SORT} links predicted positions by Kalman filter, and IoU-Tracker \cite{bochinski_AVSS2017_IoU-Tracker}  uses IoU with detection results from the previous frame. This approach tends to fail when objects move quickly or occlusions occur. The second approach is based on features \cite{luo_CVPR2019,zhang_IJCV2021_FairMOT}. Unlike IoU-based approaches, the feature-based approach links based on the similarity of features of the detected persons, making it robust to fast motion and occlusion. Yet another approach is MOTR \cite{zeng_ECCV2022_MOTR}, which is based on DETR and learns one query to continue tracking the same persons. 

Our proposed method is inspired by both MOTR and feature-based approaches;
learning to bring query features in different frames but responsible for the same person closer.


\begin{figure}[t]
    \centering
    \includegraphics[width=\linewidth]{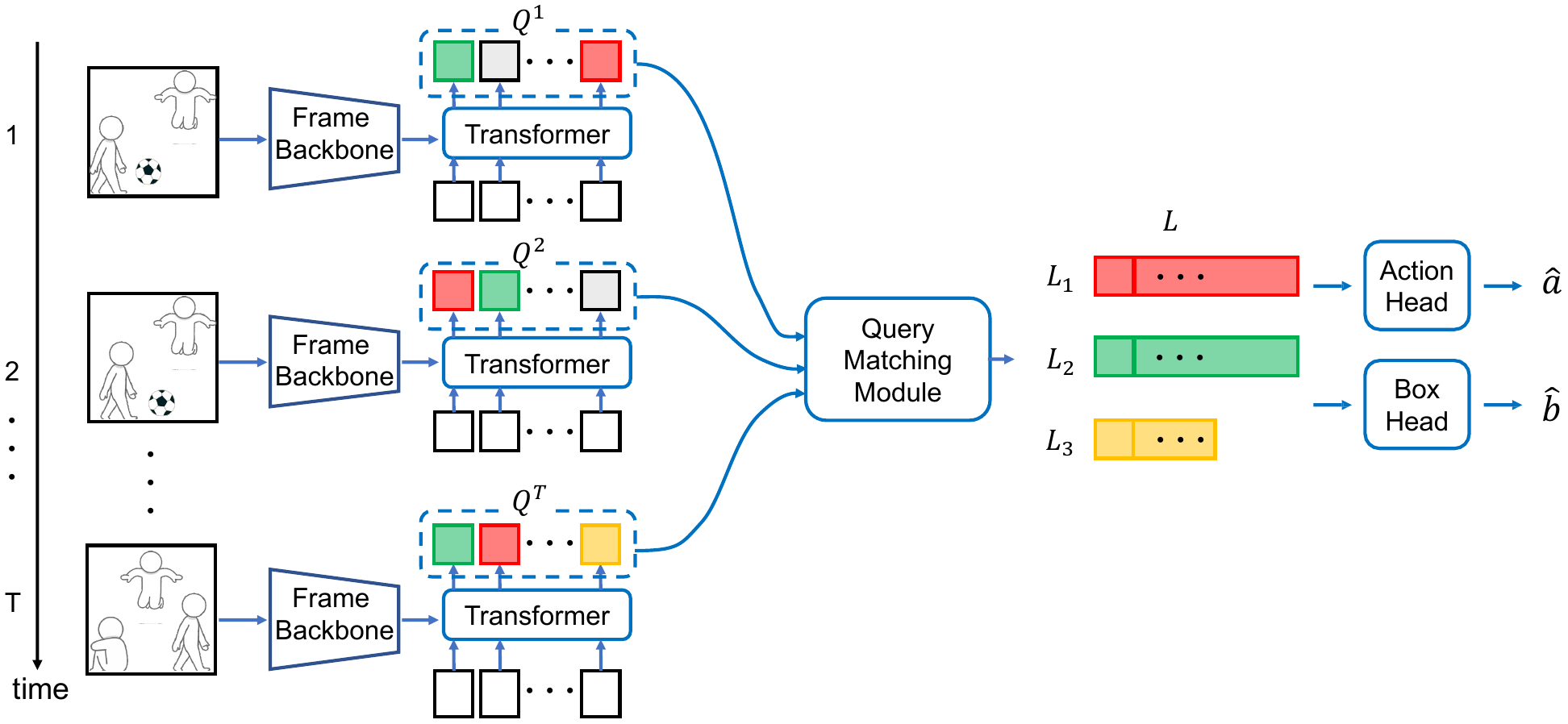}

    \caption{
        Overview of the proposed method.
        First, the frame features are obtained using the frame backbone.
        Next, the frame features and queries interact in a transformer.
        Then, the proposed Query Matching Module (QMM) matches the queries responsible for the same person in different frames.
        Finally, the output of QMM is classified by the action head to predict actions and bounding boxes in each frame.
    }
    
    \label{fig:overview}
\end{figure}

\section{Method}

In this section, we describe the proposed method that directly outputs the tube without using the IoU in the inference stage. Section \ref{sec:overview} provides an overview and presents our approach for linking the same person using queries. Section \ref{sec:qmm} explains the details of the proposed Query Matching Module (QMM), which is responsible for matching queries to the same person in different frames and construct tubes. Section \ref{sec:head} describes the details of the action head, to predict the actions of the tube.


\begin{figure*}[t]
    \centering
    \includegraphics[width=\linewidth]{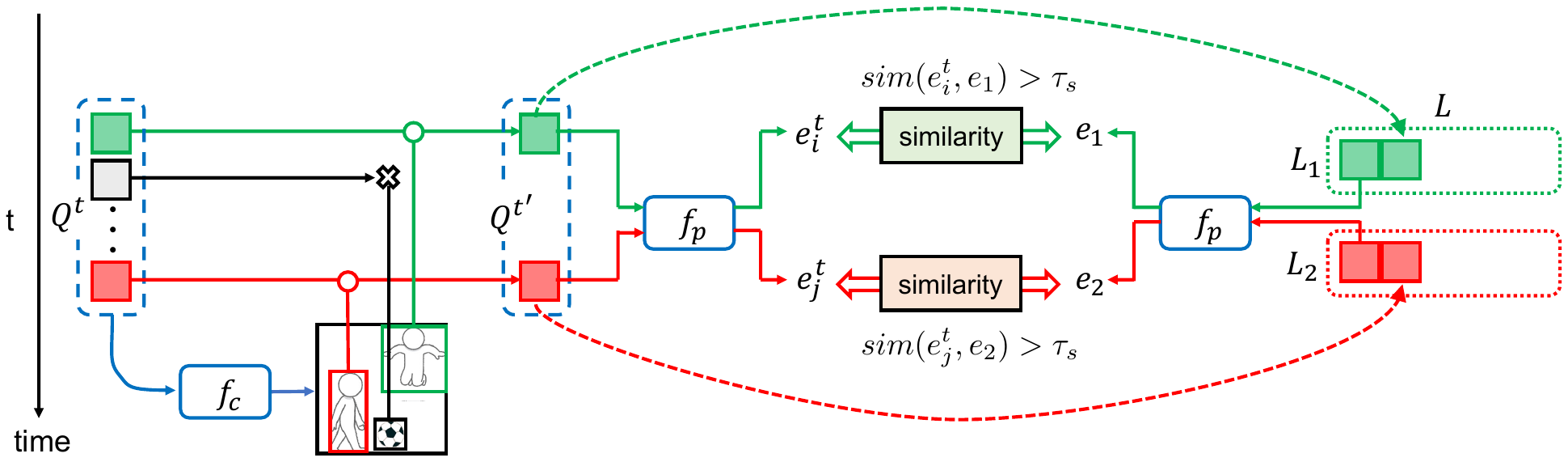}

    \caption{
Overview of the QMM Procedure.
Initially, we collect the query set $Q^t$ at frame $t$.
Subsequently, we filter out queris $Q^t$ that are not responsible for any person to form the query set $Q^{t \prime}$.
Next, we apply $f_p$ to queries in $Q^{t \prime}$ and in the list $L_j$ for the same person.
Finally, we compute the similarity between the encoded features and, if it surpasses a threshold, 
we identify it as the same person and add the  query to the corresponding list.
    }
    \label{fig:qmm_process}
\end{figure*}


\subsection{Overview}\label{sec:overview}

First, we explain the approach of linking the same person using queries. DETR \cite{Carion_ECCV2020_DETR} introduced the concept of queries and performed object detection as a bipartite matching between queries and ground-truth bounding boxes. In other words, one query is responsible for one object. The proposed method learns to match the DETR's object queries responsible for the same person between frames, which solves the drawbacks of IoU-based linking. Then, by classifying queries of the same person obtained through linking for predicting the action of the tube represented by the queries, we address the drawbacks of the use of clips.

The overview of the proposed method is shown in Figure \ref{fig:overview}. Our proposed model consists of five components: the frame backbone for extracting frame features, the transformer for interacting the frame features and the queries, the Query Matching Module (QMM) for matching queries that are responsible for the same person, the action head for predicting actions from queries of the tube, and the box head for predicting bounding boxes in the tube.

\subsubsection{Steps for prediction}

The following are the steps of how the components work.

\begin{enumerate}

\item For each frame $x^t \in \R^{3 \times H_0 \times W_0}$ for the frame index $t=1,\ldots,T$, the frame features $f^t \in \R^{3 \times H \times W}$ 
with dimensions of height $H$ and width $W$ are obtained by the frame backbone.
Here, $T$ represents the number of frames in the video,
$H_0$ and $W_0$ represent the height and width of each frame, respectively.

\item The transformer makes the frame feature $f^t$ interact with $N$ object queries to output a set of queries $Q^t = \{q_i^t \}^N_{i=1}$, where $q_i^t \in \R^d$. This process is performed for each frame $t$ to generate the sets $Q = \{Q^t \}^T_{t=1}$ for all frames $t=1,\ldots,T$.

\item QMM takes the sets $Q$ and outputs
$L_j = [q_{j^*}^{t_s}, q_{j^*}^{t_s+1}, \dots, q_{j^*}^{t_e}]$.
This is a list of queries responsible for person $j$,
starting at frame $t_s$ and ending at frame $t_e$.
Here, $q_{j^*}^t$ is a query $q_i^t \in Q^t$ for some index $i = j^*$ that is predicted to be
responsible for the same person $j$.

\item The action head takes the list $L_j$ of length $t_e - t_s + 1$ 
to predict a sequence of action scores
$\hat{a}_j = [\hat{a}_{j^*}^{t_s}, \hat{a}_{j^*}^{t_s+1}, \ldots, \hat{a}_{j^*}^{t_e}]$. 
Here,
$\hat{a}_{j^*}^t \in [0,1]^{C + 1}$ is the action score predicted at frame $t$
over $C+1$ action classes including ``no action''.
This enables us to predict different actions in different frames of the same person. 

\end{enumerate}

\subsubsection{Tube prediction}

Based on the prediction $\hat{a}_j$ obtained in the above steps,
a tube prediction is generated using the following procedure.

The action score of the tube for person $j$
is the average of the scores of the same action in $\hat{a}_j$.
This is formally defined as follows.
First, we define the following sets:
\begin{align}
    C_j &= \{ c \mid c \in \mathrm{top_k}(\hat{a}_{j^*}^t),
    \ \hat{a}_{j^*}^t \in \hat{a}_j\}
    \label{eq:topk_class} \\
    T_{j,c} &= \{ t \mid c \in \mathrm{top_k}(\hat{a}_{j^*}^t), 
    \ \hat{a}_{j^*}^t \in \hat{a}_j \}.
    \label{eq:topk_class_time}
\end{align}
Here, $\mathrm{top_k}(\hat{a}^t)$ is the operation of extracting the top $k$ elements of the score vector $\hat{a}^t$.
$C_j$ is the set of classes with top-k scores in each frame
and $T_{j,c}$ is the set of frame indices where the class $c$ is in the top-k.

The scores of the tube for each $c \in C_j$ is then defined as
\begin{align}
    \hat{a}_{j,c} = 
    \begin{cases}
        \frac{1}{|T_{j,c}|} \sum_{t \in T_{j,c}} \hat{a}_{j^*,c}^t & |T_{j,c}| > \tau_k\\
        0 & \text{otherwise},
    \end{cases}
\end{align}
where $\tau_k$ is a threshold for filtering out short predictions.

The bounding boxes of the tube for action $c$ is also defined as
\begin{align}
    \hat{b}_{j,c} &= \{ f_b(q_{j^*}^t) \mid t \in T_{j,c},
    \ q_{j^*}^t \in L_j \},
\end{align}
where $f_b: \R^d \mapsto [0,H-1] \times [0,W-1]$ is the box head that predicts the bounding box of the query.

The parameters of the pre-trained DETR are fixed and used for the frame backbone, transformer, and box head, while the QMM and action head are trained separately.
Next, the details of the QMM and the action head will be explained.

\subsection{Query Matching Module}\label{sec:qmm}

The proposed Query Matching Module (QMM) is trained using metric learning to match queries that are responsible for the same person across frames. Specifically, the query features of the same person are brought closer together, while those of different persons are pushed apart. In the following, we formulate the loss for the training, followed by the inference process.

\subsubsection{Training}

For training the matching queries of the same person, we do not use all frames of the video. Instead, we use a clip of $T$ frames extracted from the original video.

Given the sets of input queries $Q = \{Q^t \}^T_{t=1}$,
first we filter out queries to retain only those classified as ``person'' with high confidence. This is done by examining the object class score $c_i^t$
\begin{align}
     c_i^t &= f_c(q_i^t), \quad q_i^t \in Q^t,
\end{align}
obtained from $f_c: \R^d \mapsto \R^{(C+1) \times [0,1]}$, the head of class prediction of the pre-trained DETR.

Then we select the score of the person class from $c_i^t$ among all $C+1$ classes,
and queries are kept if their scores exceed the threshold $\tau_p$; otherwise, they are excluded.
\begin{align}
     Q^{t \prime} &= \{
        q_i^t \mid c_i^t[\text{``person''}] \geq \tau_p
     \}
\end{align}

Next, we find which person is responsible for each query in $Q^{t \prime}$.
The prediction of the bounding box $b_i^t = f_b(q_i^t)$ obtained using $f_b$
is compared with the ground truth box $g_j^t$ of person $j$ to compute the IoU.
We assign the person ID $j$ to only queries $q_i^t$ that exceed the threshold $\tau_{iou}$,
otherwise $-1$, as follows;
\begin{equation}
    y_i^t =
    \begin{cases}
          j, & \text{if} \ \mathrm{IoU}(b_i^t, g_j^t) > \tau_{iou}  \\
          -1, & \text{otherwise}.
    \end{cases}
    \label{eq:person_label}
\end{equation}

Next, the person feature encoder $f_p$ is used to transform the query to features $e_i^t = f_p(q_i^t)$, which are suitable for matching. These steps are performed for all frames within the clip, resulting in the set $E = \{ e_i^t \ | \ q_i^t \in Q^{t \prime} \}$
to perform contrastive learning with the following N-Pair Loss \cite{Sohn_NIPS2016_N-PairLoss};
\begin{align}
    l_i^t(\e_i^t) &= - \log
    \left(
        \frac
        {\sum_{\e_j^u\in {E_i^t}^+}
            \exp( \frac{1}{\tau_t} \mathrm{sim}(\e_i^t, \e_j^u) )}
        {\sum_{\e_k^u \in E - \{e_i^t\} }
            \exp( \frac{1}{\tau_t} \mathrm{sim}(\e_i^t, \e_k^u) )}
    \right).
    \label{eq:NPairLoss}
\end{align}
Here, 
${E_i^t}^+ = \{e_j^u \mid y_j^u = y_i^t, e_j^u \in E - \{e_i^t\} \}$
is the positive example set, that is,
the subset of $E$ where each element $e_j^u$ has the same person ID $y_i^t$ with $e_i^t$.
$\mathrm{sim}(\cdot, \cdot)$ is the cosine similarity and $\tau_t$ denotes the temperature.
We train the person feature extractor $f_p$ to minimize this contrastive loss $\mathcal{L}$,
\begin{align}
    \mathcal{L} &= \sum_{t=1}^T \frac{1}{|E|} \sum_{\e_i^t \in E} l_i^t(\e_i^t).
\end{align}


\begin{algorithm}[t]

\caption{
    Inference process of the Query Matching Module (QMM).
    Detected person queries $q^t_i$ are filtered,
    and added in lists $L$ if their scores are large enough. 
    The returned set $L$ have lists with a sufficient number of queries.
}
\label{alg:qmm}

\algrenewcommand{\algorithmicindent}{0.75em}

\begin{algorithmic}[1]
    \Require $Q = \{Q^1, \ldots, Q^T \},
        Q^t = \{q_i^t\}, \tau_p, \tau_s, \tau_k$
    \Ensure  $L = \{ L_j \}$ \Comment{a set of lists}

    \State $L \gets \emptyset$ 

    \For {$t = 1, \ldots, T$}

        \State $Q^{t \prime} \gets \emptyset$
                \Comment{init $Q^{t \prime}$ until line \ref{line:end of Qt'}}
        \For {$q_i^t \in Q^t$}
            \State $c_i^t \gets f_c(q_i^t)$
            \If {$c_i^t[\text{``person''}] \geq \tau_p$}
                \State $Q^{t \prime} \gets Q^{t \prime} \cup \{ q_i^t \}$ \label{line:end of Qt'}
            \EndIf
        \EndFor

        \If {$|Q^{t \prime}| = 0$} \Comment{No persons detected at $t$}
            \State continue
        \EndIf

        \If {$L = \emptyset$} \Comment{No persons before $t$}
            \State $L \gets \{[q_i^t] \ | \ q_i^t \in \ Q^{t \prime}\}$
                    \Comment{init with new lists}
            \State continue
        \EndIf

        \State $Q_j \gets \{ L_j[ |L_j|-1 ] \mid L_j \in L \}$ 
        \State $S \gets \{ \{ q_i^t, q_j^t, \mathrm{sim}(f_p(q_i^t), f_p(q_j^t)) \} $
        \Statex \hspace{.15\linewidth} $\mid q_i^t \in Q^{t \prime}, q_j^t \in Q_j \}$
            \Comment{a set of triplets}
        \State sort $S$ with similarity in descending order

        \For {$q_i^t, q_j^t, s_{ij}$ from $\mathrm{top_N}(S)$}
            \State Let $L_j$ be the list from which $q_j^t$ comes
            \If {already $q_i^t$ is used or $L_j$ is appended}
                \State $L \gets L \cup \{ [q^t_i] \}$ \Comment{add as a new list}
            \Else
                \If {$s_{ij} > \tau_s$}
                    \State $L_j \gets L_j + q^t_i$ \Comment{append to $L_j$}
                \Else 
                    \State $L \gets L \cup \{ [q^t_i] \}$ \Comment{add as a new list}
                \EndIf
            \EndIf
        \EndFor  
    \EndFor

\State $L \gets \{L_j \mid L_j \in L, |L_j| > \tau_k \}$

\end{algorithmic} 
\end{algorithm}


\subsubsection{Inference}

The goal of QMM during inference is to produce a set of lists $L = \{ L_j \}$ whose elements $L_j$ contain queries representing the person $j$, with $L$ initialized as $\emptyset$
as shown in Figure \ref{fig:qmm_process}.

First, as in the training phase, a set $Q^{t \prime} = \{ q_i^t \}$ is formed by selecting only queries with high person class scores to focus on those queries responsible for persons.
Then, the person feature vectors $e_i^t = f_p(q_i^t)$ are computed for each query, and also $e_j = f_p(L_j[|L_j|-1])$ for the most recent (or last) elements in the lists $L_j \in L$.
Lastly, the similarities between $e_j$ and $e_i^t$ are calculated to verify whether the current query $q_i^t$ represents the same person as $L_j$. If similarity exceeds the threshold $\tau_{s}$, the person represented by $q_i^t$ is considered the same as $L_j$, and $q_i^t$ is added to $L_j$.

If $L = \emptyset$, it is considered that no persons have appeared in the video before
and lists $L_i = [q_i^t]$ are added to $L$ for each $q_i^t \in Q^{t \prime}$.
If the similarity is below the threshold, $[q_i^t]$ is considered a newly detected person in the video, not assigned to any $L_j \in L$, and added to $L$ as a new list.

This process is iterated from $t=1$ to $T$,
and lists in $L$ with a minimum length of $\tau_k^\prime$ are used as the final output of QMM.

    
The pseudocode describing the above QMM process is shown in Algorithm \ref{alg:qmm}.


\subsection{Action Head}\label{sec:head}

The action head takes $L_j = [q_{j^*}^{t_s}, \dots , q_{j^*}^{t_e}]$, the QMM output,
and predicts the actions as $\hat{a}_j = [\hat{a}_{j^*}^{t_s}, \hat{a}_{j^*}^{t_{s+1}}, \dots \hat{a}_{j^*}^{t_e}]$.
$L_j$ represents the same person in successive frames, regardless of the presence, absence, or changes in actions. 
Therefore, the action head predicts actions $\hat{a}_{j^*}^{t} \in \mathbb{R}^{C + 1}$, including ``no action,'' for every frame $t=t_s,\ldots,t_e$ in $L_j$, rather than a single action for the entire $L_j$.

Since the length of $L_j$ varies for different $j$, a transformer is used as the action head with time encoding. This encoding adds (or concatenates) time information to encode each $q^t$ in $L_j$ with its time $t$ relative to $t_s$. The action head also has cross-attention from the frames. A pre-trained action recognition model is applied to the frames to obtain a global feature, which is then used as a query in the attention within the transformer of the action head.

Note that the action head classifies the elements in $L_j$, which are the original DETR query $q^t_i$ rather than the person feature $f_p(q^t_i)$. This choice stems from the differing goals of metric learning and action classification. Metric learning aims to keep output features distinct for different individuals, even when performing the same action. In contrast, action classification needs similar features for the same action, regardless of the person performing it. Therefore, encoded features $f_p(q^t_i)$ are employed for metric learning, while DETR queries $q^t_i$ are used for action classification.

The loss for actions is a standard cross entropy;
\begin{align}
    \mathcal{L} &= 
    \sum_{L_j \in L}
    \sum_{t=t_s}^{t_e}
    L_\mathit{CE}( \hat{a}_{j^*}^{t}, a_{j^*}^{t} ).
\end{align}
Here, action labels of each frame $t$ is
\begin{equation}
    a_{j^*}^t =
    \begin{cases}
          \text{``no action''}, & \text{if} \ y_{j^*}^t = -1  \\
          c_j^t, &  \text{otherwise},
    \end{cases}
    \label{eq:action_label}
\end{equation}
and $c_j^t$ is the action class for person $j$ at frame $t$,

\section{Experimental results}

The proposed method is evaluated using datasets commonly used for evaluating STAD.

\subsection{Settings}

\subsubsection{Datasets}

JHMDB21 \cite{Jhuang_CVPR2013_JHMDB} includes movies and YouTube videos, consisting of 928 videos with 15 to 40 frames each. There are 21 action classes, and each video contains exactly one ground truth tube of the same length as the video.

UCF101-24 \cite{Singh_ICCV2017_UCF101-24} consists of 3207 videos with clips ranging from approximately 3 to 10 seconds. There are 24 action classes, and unlike JHMDB, one video can contain any number of ground truth tubes of any length. However, all tubes within a single video belong to the same action class.

AVA \cite{Gu_CVPR2018_AVA} comprises 430 15-minute videos collected from movies. It features 80 action class labels, with 60 used for evaluation. Each video contains multiple ground truth tubes of varying lengths, allowing tubes of different action classes to coexist within a single frame.
The annotations, provided at 1-second intervals, are insufficient for training. To address this, we interpolate annotations during both training and inference using linear interpolation and object detection.
For linear interpolation, when the same person performing the same action is annotated in both the starting and ending frames of the target interval, we linearly interpolate the bounding box coordinates between these two frames.
However, linear interpolation has the drawback of large errors in bounding boxes when the person's position changes significantly. Therefore, we apply object detection to each frame using YoLoX \cite{Ge_arXiv2021_YOLOX} pre-trained on COCO \cite{Lin_ECCV2014_COCO}. We compare the Intersection over Union (IoU) between bounding boxes obtained by the YoLoX detection and linear interpolation, and use the person ID and action ID of the linear interpolated bounding boxes with the highest IoU with the detected bounding boxes.

\subsubsection{Model}

For the frame backbone, transformer, box head $f_b$ and $f_c$,
we used DETR \cite{Carion_ECCV2020_DETR} pre-trained on
COCO \cite{Lin_ECCV2014_COCO} and fixed the parameters.
However, 
we fine-tuned DETR's parameters for the UCF101-24 dataset
due to discrepancies between the detection boxes of COCO-trained DETR and the annotation boxes.
In QMM, the person feature encoder $f_p$ is a 3-layer MLP,
and the action head $f_a$ uses a two-layer transformer encoder.
The global feature used in the action head is computed by
X3D-XS \cite{Feichtenhofer_CVPR2020}, a lightweight CNN-based action recognition model,
pretrained on Kinetics \cite{kay_arXiv2017}.

\begin{table*}[t]
    \centering
    \caption{
        Comparison of recall performance between QMM and IoU-based linking.
        The 3D IoU threshold is set at 0.5 for JHMDB and 0.2 for both UCF101-24 and AVA.
    }
    \label{tab:qmm}

    \begin{tabular}{l@{\hskip 10pt}c@{ }c@{ }c@{ }c@{}c@{\hskip 10pt}c@{ }c@{ }c@{ }c@{ }c@{\hskip 10pt}c@{ }c@{ }c@{ }c}
                        & \multicolumn{4}{c}{JHMDB} && \multicolumn{4}{c}{UCF} && \multicolumn{4}{c}{AVA} \\ \cline{2-5} \cline{7-10} \cline{12-15}
                        & All        & L          & M     & S         && All       & L         & M         & S         && All     & L       & M       & S       \\ \hline
        IoU $\geq$ 0.25 & \tb{92.9}  & \tb{75.8}  & 73.2  & \tb{92.3} && 82.9      & 57.4      & \tb{55.5} & 68.0      && 87.7  & 25.8  & 20.0  & 81.3  \\
        IoU $\geq$ 0.5  & 91.0       & 63.6       & 73.2  & 91.8      && 78.6      & 44.4      & 49.6      & \tb{71.1} && 90.9  & 21.7  & 16.7  & 86.9  \\
        IoU $\geq$ 0.75 & 81.0       & 27.3       & 53.7  & 88.7      && 50.7      & 4.00      & 21.2      & 68.0      && \tb{93.0}  & 14.1 & 12.9 & \tb{92.3} \\
        QMM             & 91.4       & \tb{75.8}  & \tb{75.6} & 91.2  && \tb{83.3} & \tb{60.1} & 53.9      & 68.7      && 83.9 & \tb{37.8} & \tb{23.0} & 69.1 \\
\end{tabular}

\end{table*}

\subsubsection{Training and inference strategies}

QMM and action head are trained separately. First, QMM is trained, followed by the training of the action head.

QMM is trained for a clip of eight frames extracted at 4-frame intervals from a random start time in the video for JHMDB and UCF101-24, while for AVA, a randomly selected key frame is used as the starting frame. Each frame is resized while maintaining its aspect ratio so that the longer side becomes 512 pixels, which is the input size of the model. The shorter side is padded with black to create the clip. The training settings are as follows: 20 epochs, batch size of 8, AdamW optimizer with an initial learning rate of 1e-4 (decayed by the factor of 10 at epoch 10). Furthermore, $\tau_{iou}=0.2, \tau_t=1$, and $\tau_p=0.75$ for JHMDB, $\tau_p=0.5$ for UCF101-24 and AVA.

The action head is trained using the output from QMM inference. During inference, frame-by-frame preprocessing remains the same as in the training phase. For JHMDB and UCF101-24, all frames are used as input. However, for AVA, due to memory constraints, the number of input frames during inference is limited to 16. The parameter settings are as follows: For JHMDB, $\tau_p=0.9$, $\tau_s=0.5$, $\tau_k^\prime=8$,
for UCF101-24, $\tau_p=0.5$, $\tau_s=0.25$, $\tau_k^\prime=16$, and
for AVA, $\tau_p=0.5, \tau_s=0.25, \tau_k^\prime=8$. The training settings are as follows: 20 epochs, AdamW optimizer with an initial learning rate of 1e-3  (decayed by the factor of 10 at epoch 15). When generating tubes, $\tau_{k}=8$ is used.

\subsubsection{Metrics}

For JHMDB and UCF101-24, we adopt video-mAP (v-mAP) as the most commonly used evaluation metric for STAD. This is the class average of average precision based on the spatio-temporal IoU (3D IoU) between predicted tubes and ground truth tubes for each action class. On the other hand, for AVA, we use frame-mAP on annotated key frames as the evaluation metric, as it is the official protocol.

We also evaluated tube detection solely to confirm the effectiveness of the proposed QMM. Specifically, we evaluate using the recall of predicted tubes against the regions of ground truth tubes. We consider a prediction correct if the 3D IoU between the ground truth tube and the predicted tube is above a threshold and incorrect otherwise; then calculate the recall. Since the STAD dataset only annotates people performing actions, we use only recall for evaluation rather than precision, which is affected by other people who are not performing any actions. For the same reason, we use the 3D IoU only at frames where the ground truth tube exists when calculating the recall.

Furthermore, we perform evaluations based on the magnitude of action position changes. We adopt the motion category in MotionAP \cite{Singh_WACV2023_TAAD}, and calculate the recall for ground truth tube regions in each of the Large (L), Medium (M), and Small (S) motion categories.

\subsection{Results}

\subsubsection{Performance of query matching by QMM}
\label{sec:ex_qmm}

Table \ref{tab:qmm} shows a comparison of the proposed QMM versus IoU-based linking. In the table, ``All'' represents the recall for all ground truth tubes regardless of motion category, while ``L,'' ``M,'' and ``S'' represent the recall for each motion category. The 3D IoU threshold is set to 0.5 for JHMDB and 0.2 for UCF101-24 and AVA.

The performance of IoU-based linking decreases for categories with larger motions as the threshold increases for both datasets, indicating that using a smaller IoU threshold value could mitigate the drawbacks of IoU-based linking. However, using a small threshold value leads to fundamental problems such as easier linking with false detections and increased possibility of linking failures in situations where people overlap.

In contrast, QMM, compared to using an IoU threshold of 0.75, the proposed QMM performs better than IoU-based linking with an IoU threshold of 0.75. Compared to an IoU threshold of 0.5, it improves in the All, Large, and Medium categories, while performance decreases in the Small category. Compared to an IoU threshold of 0.25, for JHMDB and AVA, performance decreases in the All and Small categories but is equal to or better in the Large and Medium categories. However, for UCF101-24, performance improvements are observed in the All, Large, and Small categories. From these findings, QMM is particularly effective for actions with large position changes.

\begin{table}[t]
    \centering
    \caption{
        Ablation study on the use of time encoding and global features in the action head.
        Performance is shown as v-mAP@0.5 for JHMDB, and v-mAP@0.2 for UCF101-24.
    }
    \label{tab:v-mAP}

    \begin{tabular}{@{}c@{}c@{}c@{ }c@{ }c@{ }c}
            &        & \multicolumn{2}{c}{JHMDB} & \multicolumn{2}{c}{UCF}  \\ \cline{3-6}
        \begin{tabular}{c} time \\ encoding \end{tabular} & \begin{tabular}{c} global \\ feature \end{tabular}                                  & top1 & top5 & top1 & top5 \\ \hline
         -  &               & 28.7 & 39.0 & 42.4 & 49.4 \\
        add &               & 31.9 & 42.1 & 44.0 & 50.1 \\
        concat &            & 35.7 & 46.5 & 45.0 & 51.1 \\
        concat & \checkmark & 41.7 & 53.3 & 61.0 & 65.6 \\
    \end{tabular}

\end{table}

\subsubsection{Ablation study of action head}\label{sec:ex_head}

Here, we show an ablation study on the action head. Specifically, we compare the presence and absence of time encoding, which adds temporal information to the input, and global features obtained from a pre-trained model. Table \ref{tab:v-mAP} shows the results based on v-mAP. Note that top1 and top5 represent the top-k values in Eqs \eqref{eq:topk_class} and \eqref{eq:topk_class_time}, respectively. The 3D IoU threshold is set to 0.5 for JHMDB and 0.2 for UCF101-24, consistent with the previous experiment. In other words, the performance is shown as v-mAP@0.5 for JHMDB and v-mAP@0.2 for UCF101-24.
%

Using the time encoding was effective in both datasets. In particular, concatenation showed a more pronounced effect than addition. This suggests that explicit incorporation of temporal information is crucial for effective action recognition.

Next, we can see that incorporating global features from a pre-trained action recognition model significantly improves performance in both datasets. This improvement leads to two key insights.
First, global features support the learning of the action head, which needs to be trained on the output queries of the pre-trained DETR. Using these global features for cross-attention in the action head, the head is likely to achieve more effective learning from scratch.
Second, using DETR's output, which is trained on object detection tasks, for action recognition has limitations. In object detection, all people are labeled as ``person,'' regardless of their actions. As a result, DETR's queries may lack action-specific information. However, global features can fill this gap, boosting action recognition performance.

\begin{figure*}[t]
    \centering
    \includegraphics[width=.8\linewidth]{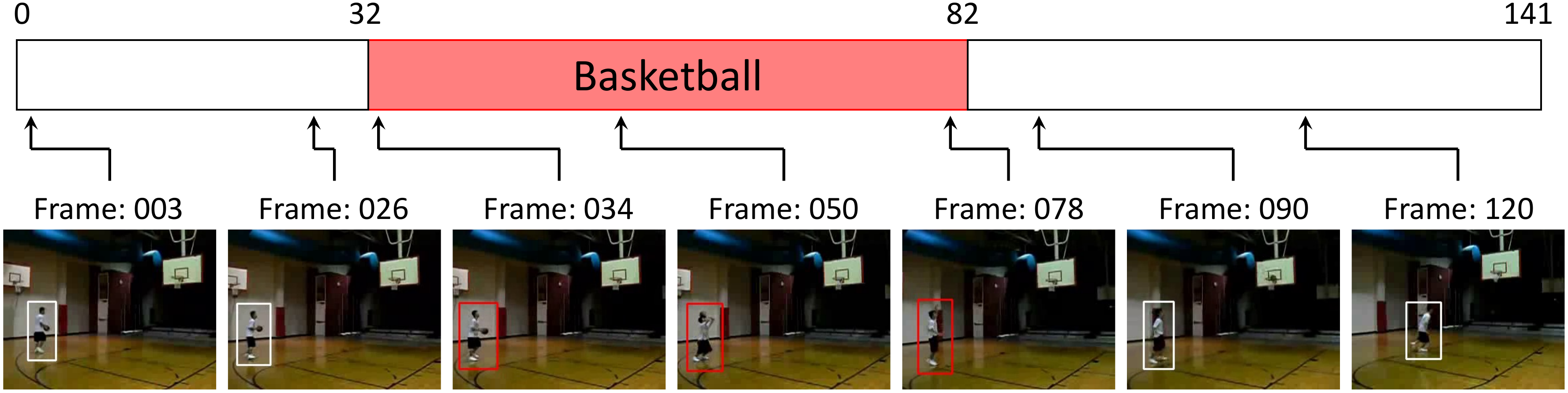}

    \caption{
    Visualization of predictions by the proposed method.
    (Top) Temporal annotation intervals.
    (Bottom) The bounding boxes represent predicted tubes.
    Boxes are in when the top-1 class of the frame is ``no action'',
    and in red when `Basketball''.
    }
    \label{fig:example}
\end{figure*}

The differences in the top-k value show that top-5 performs better than top-1. This suggests that while the top-1 action in $\hat{a}^t$ may not be the correct action of the tube, it is often the case that the top-5 actions include it. Therefore, increasing the top-k value improves the performance of the v-mAP.
However, for videos where the start and end of actions are correctly detected using only the top-1 action, as shown in Figure \ref{fig:example}, using the top-5 actions may negatively affect performance due to unnecessary actions being also used to predict the action of the tube.
Future work includes improvements such as excluding timestamps from $T_{j,c}$ where the predicted action scores are lower than the ``no action'' class score.


\begin{table*}[t]
    \centering
    \caption{
        Comparisons with state-of-the-art methods using v-mAP.
        Performance is shown as v-mAP@0.5 for JHMDB , v-mAP@0.2 for UCF101-24 and frame-mAP@0.5 for AVA.
    }
    \label{tab:sota}

  \begin{tabular}{l|cccc}
        Model               & outputs & JHMDB  & UCF101-24 & AVA  \\ \hline
        ActionTubes \cite{Gkioxari_CVPR2015_Tube}       & bbox    & 53.3    & -        & -      \\
        STMH \cite{Weinzaepfel_ICCV2015}                & bbox    & 60.7    & 46.8     & -      \\
        Saha \et \cite{saha_arXiv2016_OJLA}             & bbox    & 71.5    & 66.8     & -      \\
        ROAD \cite{Singh_ICCV2017_UCF101-24}            & bbox    & 72.0    & 73.5     & -      \\
        ACRN \cite{Sun_ECCV2018_ACRN}                   & bbox    & 80.1    & -        & 17.4      \\
        YOWO \cite{köpüklü_arXiv2021_YOWO}              & bbox    & 85.9    & 78.6     & 20.2      \\ \hline
        ACT \cite{Kalogeiton_ICCV2017_ACT-Detector}     & tubelet & 73.7    & 77.2     & -      \\
        STEP \cite{Yang_CVPR2019_STEP}                  & tubelet & -       & 76.6     & 18.6      \\
        MOC \cite{Yixuan_ECCV2020}                      & tubelet & 77.2    & 82.8     & -      \\
        TubeR \cite{Zhao_CVPR2022_TubeR}                & tubelet & 80.7    & 85.3     & 33.6      \\
        STAR \cite{gritsenko_arXiv2023_STAR}            & tubelet &\tb{92.6}&\tb{88.0} & 42.5      \\ \hline
        T-CNN \cite{Hou_ICCV2017_T-CNN}                 & tube    & 76.9    & 47.1     & -      \\
        TAAD \cite{Singh_WACV2023_TAAD}                 & tube    &  -      & 79.6     & -      \\
        ours                                            & tube    & 53.3    & 65.6     & -      \\
  \end{tabular}

\end{table*}


\subsubsection{Comparisons with other methods}
\label{sec:ex_comparison}

Table \ref{tab:sota} shows a comparison with state-of-the-art methods. In the table, "outputs" represents the types of model outputs, and performance is shown as v-mAP@0.5 for JHMDB, v-mAP@0.2 for UCF101-24, and frame-mAP@0.5 for AVA. 

Clearly, the research direction is shifting from methods that output bounding boxes to tubelets and further to tubes, which is expected to lead to performance improvements. However, methods that output tubes are still in their early stages of development and few in number, leaving room for further improvement.

Although the proposed method underperforms existing methods in the category of tube output, it demonstrates superior computational efficiency and lower resource requirements.

Many existing methods use large-scale datasets or backbone models. For example, STAR, which achieves the highest performance, uses CLIP \cite{Radford_ICML2021_CLIP} and Kinetics 700 \cite{Carreira_arXiv2022_K700} for pre-training, and employs ViViT-L \cite{Arnab_ICCV2021_ViViT}, with parameters of several hundred MB, as the backbone. Although these methods demonstrate high performance, they require enormous computational costs and resources.

In contrast, the proposed method uses only COCO pre-trained DETR, a 3-layer MLP for QMM, a 2-layer transformer for the action head, and X3D XS \cite{Feichtenhofer_CVPR2020}, the smallest version of X3D with only 11.9 MB parameters, for the global features. This lightweight approach gives the proposed method advantages in computational efficiency and low resource requirements.

Furthermore, considering the significant performance improvement achieved by using the global feature, the proposed method has the potential to improve while maintaining a balance between performance and efficiency by using larger models. This is particularly advantageous in resource-constrained environments or applications requiring real-time processing.


\section{Conclusions}

We have proposed a STAD method that directly generates action tubes without relying on post-processing steps, such as IoU-based linking. Instead, our approach utilizes metric learning between DETR queries to link actions across frames. Although the proposed Query Matching Module (QMM) and action head are trained using fixed-length clips, our method offers a flexible inference framework that can handle variable-length video inputs, enhancing its adaptability to diverse videos. In future work, our aim is to extend this approach to more complex tasks, such as spatio-temporal sentence grounding \cite{Yang_TubeDETR_CVPR2022}, and to further refine the architecture and learning strategies to improve the performance of query matching to track the same person across different frames.

\section*{\uppercase{Acknowledgments}}

This work was supported in part by JSPS KAKENHI Grant Number JP22K12090.

\bibliographystyle{apalike}
{\small
\bibliography{mybib,all}}

\end{document}